\documentclass{bmvc2k}

\usepackage{times}
\usepackage{epsfig}
\usepackage{graphicx}
\usepackage{amsmath}
\usepackage{amssymb}
\usepackage{array}
\usepackage{multirow}
\usepackage{epstopdf}
\usepackage{algorithm}
\usepackage{algpseudocode}
\usepackage[abs]{overpic}
\newcommand{\todo}[1]{{\textcolor{black}{#1}}}
\newcommand{\cui}[1]{{\textcolor{black}{#1}}}
\newcommand{\cuif}[1]{{\textcolor{black}{#1}}}

\newcommand{\figref}[1]{Figure~\ref{#1}}
\newcommand{\equref}[1]{Equation~\ref{#1}}
\newcommand{\secref}[1]{Section~\ref{#1}}
\newcommand{\tabref}[1]{Table~\ref{#1}}

\newcommand{\alref}[1]{Algorithm~\ref{#1}}

\newcommand{\ve}[1]{\mbox{{\bf #1}}} 
\newcommand{\best}[1]{{\bf #1}}

\title{Linear Global Translation Estimation \\with Feature Tracks}

\addauthor{Zhaopeng Cui}{zhpcui@gmail.com}{1}
\addauthor{Nianjuan Jiang}{nianjuan.jiang@adsc.com.sg}{2}
\addauthor{Chengzhou Tang}{chengzhout@gmail.com}{1}
\addauthor{Ping Tan}{pingtan@sfu.ca}{1}

\addinstitution{
 GrUVi Lab\\
 Simon Fraser University\\
 Burnaby, Canada
}
\addinstitution{
 Advanced Digital Sciences Center of\\
 Illinois, Singapore
}

\runninghead{Cui \bmvaEtAl}{Linear Global Translation Estimation}

\def\eg{\emph{e.g}\bmvaOneDot}

\def\etal{\emph{et al}\bmvaOneDot}
\def\ie{\emph{i.e}\bmvaOneDot}

\begin{document}

\maketitle

\begin{abstract}
This paper derives a novel linear position constraint for cameras seeing a common scene point,
which leads to a direct linear method for global camera translation estimation.
Unlike previous solutions, this method deals with \cui{collinear camera motion and weak image association} at the same time.
The final linear formulation does not involve the coordinates of scene points,
which makes it efficient even for large scale data.
We solve the linear equation based on $L_1$ norm, which makes our system more robust to outliers in essential matrices and feature correspondences.
We experiment this method on both sequentially captured images and unordered Internet images. The experiments demonstrate its strength in robustness, accuracy, and efficiency.
\end{abstract}

\section{Introduction}
\label{sec:intro}

Structure-from-motion (SfM) algorithms aim to estimate scene structure and camera motion from multiple images, and they can be broadly divided into incremental and global methods according to their ways to register cameras. Incremental methods  register cameras one by one \cite{snavely2006, wu2013visualsfm} or iteratively merge partial reconstructions \cite{fitzgibbon1998,Lhuillier2005}. These methods require frequent intermediate bundle adjustment (BA) to ensure correct reconstruction, which is computationally expensive. Yet, their results often suffer from large drifting errors. In comparison, global methods (\eg \cite{Govindu01c,MartinecP07,Crandall2011,jiang2013g,Wilson2014,Ozyesil2015}) register all cameras simultaneously,
which has better potential in both efficiency and accuracy.

\begin{figure} \centering

\begin{overpic}[width=0.9\textwidth,tics=10]
{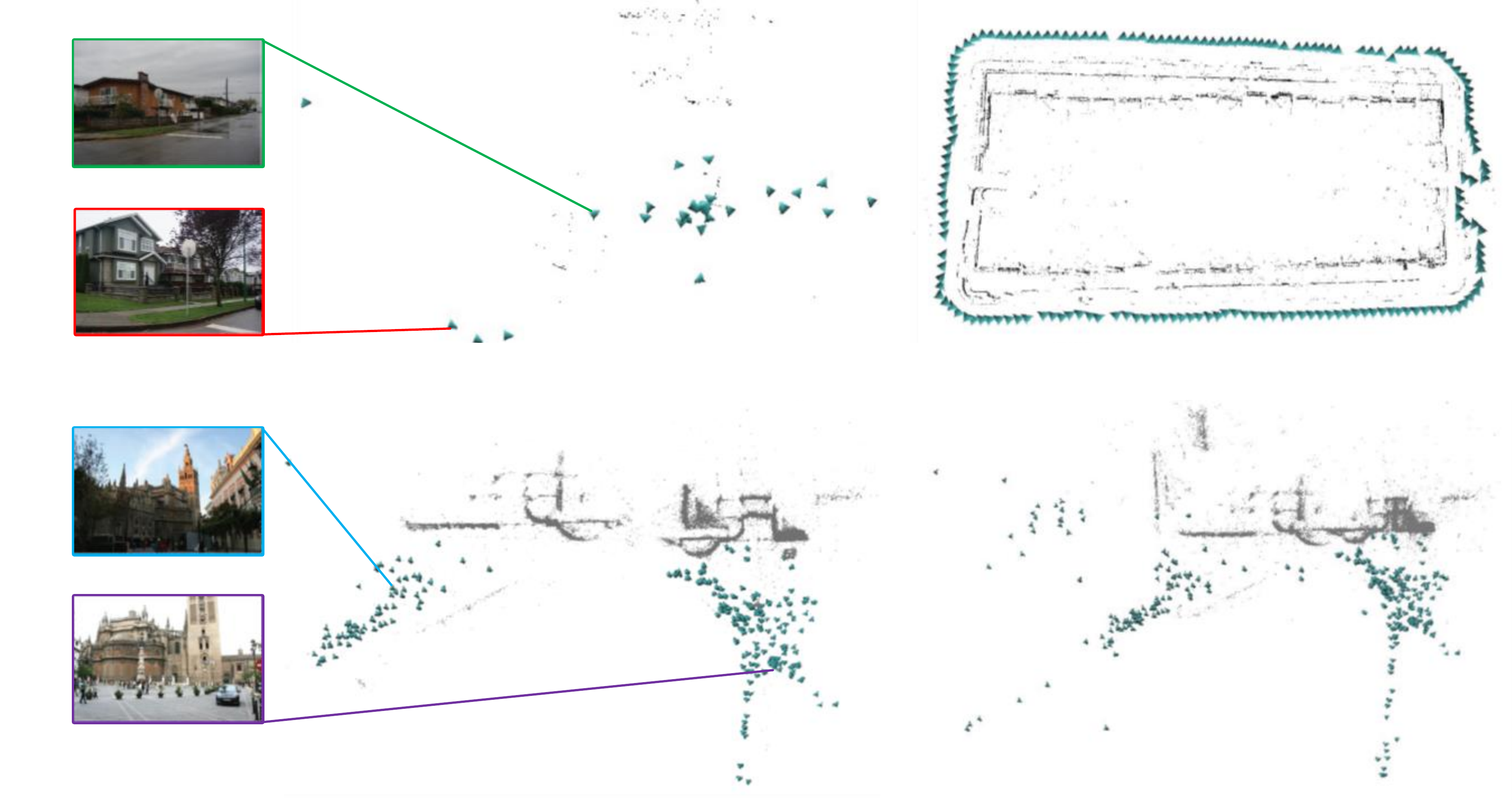}

\put(5,125){\rotatebox{90}{\small \emph{Street}}}
\put(5,40){\rotatebox{90}{\small \emph{Seville}}}

\put(100,90){\small Result from \cite{Wilson2014}}
\put(250,90){\small Our result}

\put(100,-6){\small Result from \cite{jiang2013g}}
\put(250,-6){\small Our result}

\end{overpic}
\caption{1DSFM \cite{Wilson2014} and triplet-based methods (\eg \cite{jiang2013g}) require strong association among images. 
As shown in the left, they fail for images with weak association. In comparison, as shown in the right, the results from our method do not suffer from such problems. }
\label{fig:motivation}

\end{figure}

Global SfM methods often solve the camera orientations and positions separately. The global position estimation is more challenging than the orientation estimation due to the noisy pairwise translation encoded in essential matrices \cite{enqvist11}. This paper focuses on the problem of global position (\ie translation) estimation.

Essential matrix based translation estimation methods \cite{Govindu01c, Brand04s,Arie2012} can only determine camera positions in a parallel rigid graph  \cite{Ozyesil2015}, and they usually degenerate at collinear camera motion because the translation scale is not determined by an essential matrix. Trifocal tensor based methods \cite{Sim2006,CourchayDKS10,moulon2013gf,jiang2013g} could deal with collinear motion as the relative scales are encoded in a trifocal tensor. 
However, these methods usually rely on a strongly connected camera-triplet graph, where two triplets are connected by their common edge.
The 3D reconstruction will distort or break into disconnected components when such strong association among images does not exist. By solving cameras and scene points together, some global methods \cite{Rother2003,Kahl2007m,MartinecP07,Crandall2011,sinha2010ml} can deal with collinear motion. These methods usually need to filter epipolar geometries (EGs) carefully to avoid outliers. \cui{Including scene points in the formulation also hurts the scalability of the algorithm, since there are many more scene points than cameras.} The recent 1DSfM method \cite{Wilson2014} designs a smart filter to discard outlier essential matrices and solves scene points and cameras together \cui{by enforcing orientation consistency}. However, this method requires abundant association between input images, \eg $\sim O(n^2)$ essential matrices for $n$ cameras, which is more suitable for Internet images and often fails on sequentially captured data.

The data association problem of 1DSfM \cite{Wilson2014} and triplet-based methods (here, we take \cite{jiang2013g} as an example) is exemplified in  \figref{fig:motivation}. The \emph{Street} example on the top is a sequential data. 1DSfM fails on this example due to insufficient image association, since each image is only matched to its 4 neighbors at the most. In the \emph{Seville} example on the bottom, those Internet images are mostly captured from two viewpoints (see the two representative sample images) with weak affinity between images at different viewpoints. This weak data association causes seriously distorted reconstruction for the triplet-based method in \cite{jiang2013g}.

This paper introduces a direct linear algorithm to address the presented challenges.
It avoids degeneracy at collinear motion and deals with weakly associated data. 
Our method capitalizes on constraints from essential matrices and feature tracks. 
For a scene point visible in multiple (at least three) cameras,
we consider triangles formed by this scene point and two camera centers.
We first generalize the camera-triplet based position constraint in \cite{jiang2013g} to 
our triangles with scene points.
We then eliminate the scene point from these constraints.
In this way, we obtain a novel linear equation for
the positions of cameras linked by a feature track. 
Solving these linear equations from many feature tracks simultaneously register all cameras in a global coordinate system.

This direct linear method minimizes a geometric error,
which is the Euclidean distance between the scene point to its corresponding view rays.
It is more robust than the linear constraint developed in \cite{Rother2003}, which minimizes an algebraic error. A key finding in this paper is that, a direct linear solution (without involving scene points)
exists by minimizing the \todo{point-to-ray error} instead of the reprojection error.
Since the \todo{point-to-ray} error approximates the reprojection error well when cameras are calibrated,
our method is a good linear initialization for the final nonlinear BA.

At the same time, this direct linear formulation lends us sophisticated optimization tools, such as $L_1$ norm optimization \cite{candes2005decoding, goldstein2009split, Boyd2011ADMM,parikh2013proximal}.
We minimize the $L_1$ norm when solving the linear equation of camera positions.
In this way, our method can tolerate a larger amount of outliers in both essential matrices and feature correspondences.
The involved $L_1$ optimization is nontrivial.
We derive a linearization of the alternating direction method of multipliers algorithm \cite{Boyd2011ADMM} to address it.
%


\section{Related Work} \label{sec:related_work}

\textbf{Incremental approaches.} Most of well-known SfM systems register cameras sequentially \cite{Pollefeys2004,snavely2006,snavely2008modeling, pollefeys2008detailed, agarwal2009} or hierarchically \cite{fitzgibbon1998,Lhuillier2005,havlena2009randomized} from pairwise relative motions. In order to minimize error accumulation, frequent intermediate bundle adjustment is required for both types of methods, which significantly reduces computation efficiency. The performance of sequential methods relies heavily on the choice of the initial image pair and the order of subsequent image additions \cite{haner2012covariance}.

\textbf{Global rotation estimation.} Global SfM methods solve all camera poses simultaneously. Most of these methods take two steps. Typically, they solve camera orientations first and then positions. The orientation estimation is well studied with an elegant rotation averaging algorithm presented in \cite{hartley2013rotation}. The basic idea was first introduced by Govindu \cite{Govindu01c}, and then developed in several following works \cite{Govindu04, MartinecP07, hartley2013rotation}. In particular, \cite{chatterjee2013efficient} introduced a  robust $L_1$ method which was adopted in several recent works \cite{Wilson2014,Ozyesil2015}.

\textbf{Global translation estimation.}
The translation estimation is more challenging. 
Some pioneer works \cite{Govindu01c, Brand04s, Govindu06, Arie2012} solved camera positions solely from constraints in essential matrices. 
Typically, they enforce consistency between pairwise camera translation directions and those encoded in essential matrices. 
Recently, \"{O}zyesil and Singer \cite{Ozyesil2015} prove that essential matrices only determine camera positions in a parallel rigid graph, and present a convex optimization algorithm to solve this problem. In general, all these essential matrix based methods degenerate at collinear motion, where cameras are not in a parallel rigid graph.

This degeneracy can be avoided by exploiting relative motion constraints from camera triplets \cite{Sim2006, CourchayDKS10, sinha2010ml, moulon2013gf}, as the trifocal tensor encodes the relative scale information. Recently,
Jiang \textit{et al.} \cite{jiang2013g} derived a novel linear constraint in a camera triplet and solved all cameras positions in a least square sense. While triplet-based methods avoid degenerated camera motion, they often require strong association among images -- a connected triplet graph, 
where camera triplets are connected by common edges.


Some global methods estimate cameras and scene points together. Rother \cite{Rother2003} \cui{solved camera positions and points by minimizing an algebraic error}.
Some works \cite{Kahl2007m, MartinecP07, Olsson07,AgarwalSS08,Li2009efficient} solved the problem by minimizing the $L_{\infty}$ norm of reprojection error. However, the $L_{\infty}$ norm is known to be sensitive to outliers and careful outlier removal is necessary \cite{OlssonEH10,Dalalyan09l1p}. Recently, Wilson and Snavely \cite{Wilson2014} directly solved cameras and points by Ceres Solver \cite{ceres-solver} after applying a smart filter to essential matrices. Generally speaking, involving scene points improves the robustness/accuracy of camera registration, but also significantly increases the problem scale. Feature correspondence outliers also pose a significant challenge for these methods.

Our method capitalizes on constraints in essential matrices and feature tracks. 
It avoids degeneracy at collinear motion, handles weak data association, and is robust to feature correspondence outliers. 


\section{Global Translation Estimation} \label{sec:global_motion}
Given an essential matrix between two images $i, j$ (\eg computed by the five-point algorithm \cite{nister04, Li2006}), we obtain the relative rotation $\ve{R}_{ij}$ and translation direction $\ve{t}_{ij}$ between the two cameras. Here, $\ve{R}_{ij}$ is a $3\times3$ orthonormal matrix and $\ve{t}_{ij}$ is a $3\times1$ unit vector. We further denote the global orientation and position of the $i$-th ($1\leq i \leq N$) camera as $\ve{R}_i$ and $\ve{c}_i$ respectively. These camera poses are constrained by the following equations
\vspace{-0.02in}
\begin{equation}
\vspace{-0.02in}
\begin{aligned}
\ve{R}_j = \ve{R}_{ij} \ve{R}_i,  \ \ \ \ \ve{R}_j(\ve{c}_i-\ve{c}_j) \simeq \ve{t}_{ij}.
\end{aligned}
\end{equation}
Here, $\simeq$ means equal up to a scale.

Like most global methods, we compute camera orientations first and solve camera positions after that. 
\cuif{We adopt the global rotation estimation method in \cite{chatterjee2013efficient}. In order to enhance robustness, we adopt additional loop verifications  \cite{Zach2010} on the input pairwise relative camera rotations beforehand.}
Specifically, we chain the relative rotations along a three-camera loop as $\hat{\ve{R}} = \ve{R}_{ij}\ve{R}_{jk}\ve{R}_{ki}$, and compute the angular difference \cite{hartley2013rotation} between $\hat{\ve{R}}$ and the identity matrix. If the difference is larger than a threshold $\varphi_1$ (\cui{$3$} or $5$ degrees for sequential data or unordered Internet data), we consider the verification fails. We discard an EG if every verification it participates in fails.

The key challenge in translation estimation is that an essential matrix does not tell the scale of translation. 
We seek to obtain linear equations for those unknown scales without resorting to camera-triplets.
Our translation estimation is based on a linear 
position constraint arising from a triangle formed by two camera positions and a scene point.
With this constraint, the positions of cameras linked by a feature point should satisfy a linear equation.




\begin{figure} \centering
\begin{tabular} {cc}
\includegraphics[width=0.35\linewidth]{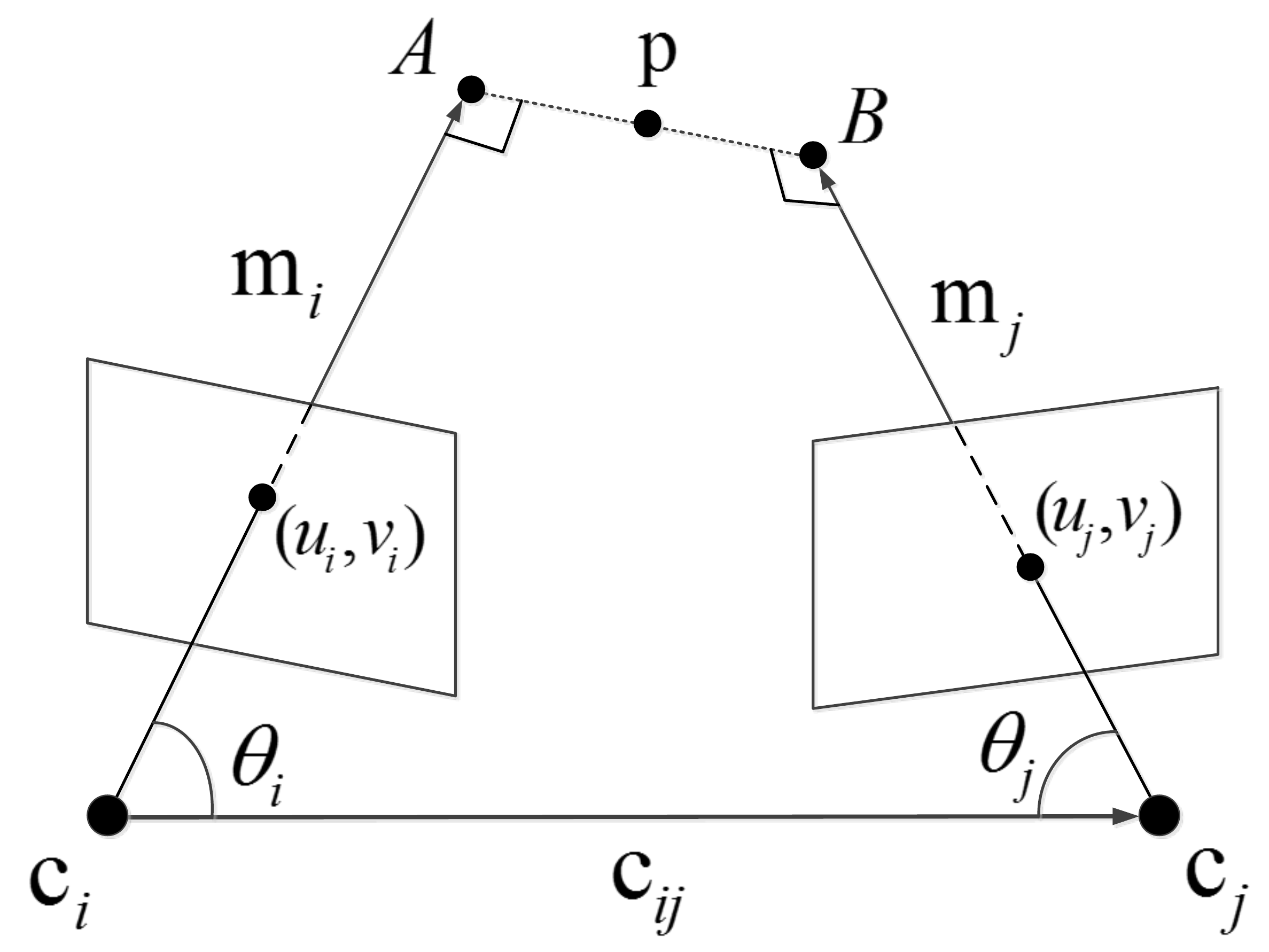} & \includegraphics[width=0.35\linewidth]{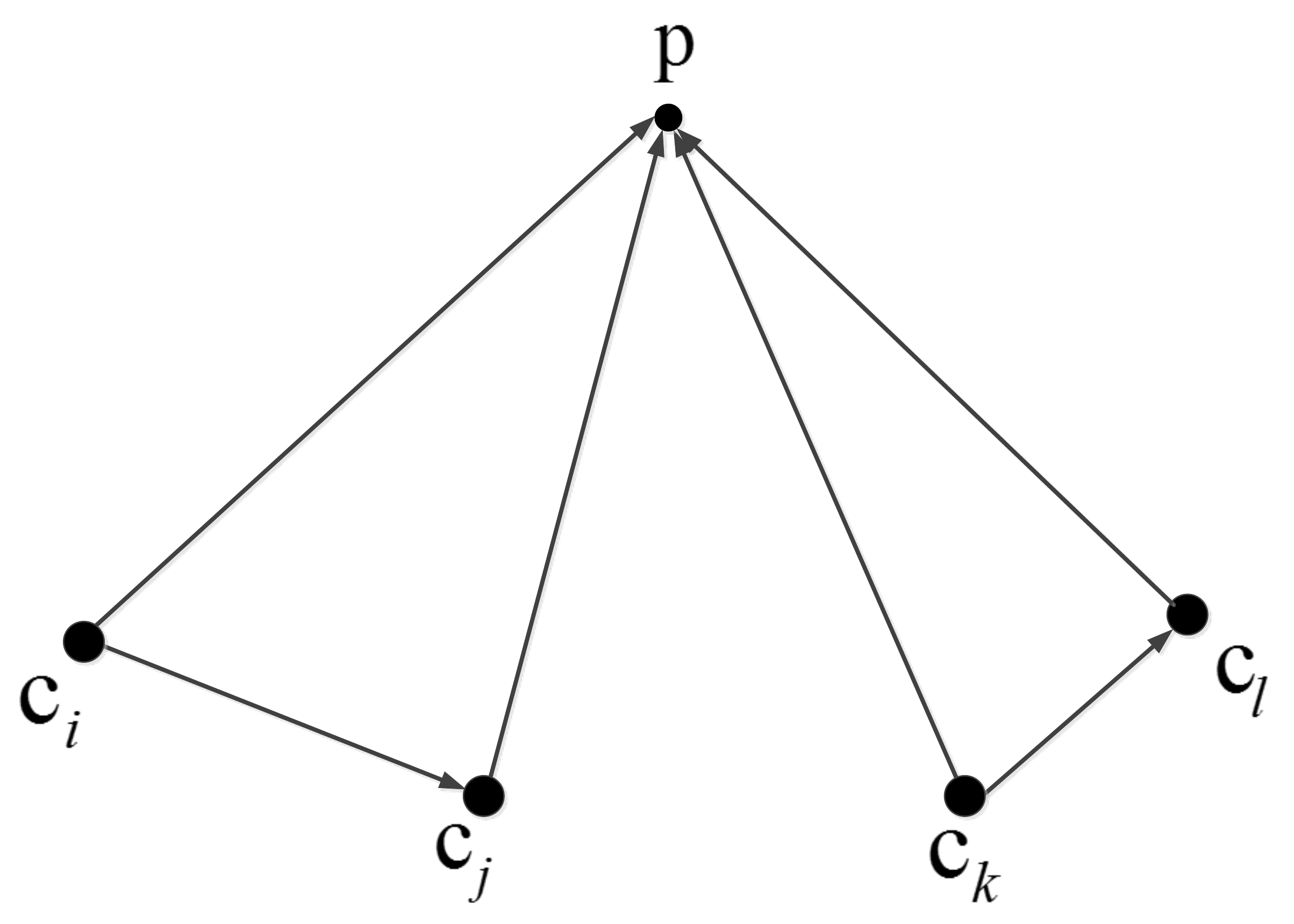}\\
(a) &(b)\\

\end{tabular}
\caption{(a) The positions of a scene point $\ve{p}$ and two camera centers $\ve{c}_i$ and $\ve{c}_j$ satisfy a linear constraint detailed in \secref{sec:triangle_constraint}. (b) The positions of four cameras seeing the same scene point satisfy a linear constraint detailed in \secref{sec:feature_constraint}. }
\label{fig:triangulation}
\vspace{-0.05in}
\end{figure}

\subsection{Constraints from a triangle} \label{sec:triangle_constraint}



A linear constraint on positions of cameras in a triplet is derived in \cite{jiang2013g}. 
We generalize it to the case of triangles formed by a scene point and two cameras. 
As shown in \figref{fig:triangulation}, we compute the location of a scene point $\ve{p}$ as the middle point of the mutual perpendicular line segment
$AB$ of the two rays passing through $\ve{p}$'s image projections. Specifically, it is computed as
\vspace{-0.02in}
\begin{equation} \label{equ:twoViewT_a}
\vspace{-0.02in}
\ve{p} = \frac{1}{2}(A+B) = \frac{1}{2}(\ve{c}_i + s_{i}\ve{m}_{i} + \ve{c}_j + s_{j}\ve{m}_{j}).
\end{equation}
Here, $\ve{c}_i$ and $\ve{c}_j$ are the two camera centers. The two unit vectors $\ve{m}_{i}$ and $\ve{m}_{j}$ origin from the camera centers and point toward the image projections of $\ve{p}$. $s_i$ and $s_j$ are the distances from the points $A, B$ to $\ve{c}_i, \ve{c}_j$ respectively, \ie $A = \ve{c}_i +s_i \ve{m}_i$ and $B = \ve{c}_j + s_j \ve{m}_j$.

\textbf{The rotation trick.} \cuif{ The rotation trick used in \cite{jiang2013g} shows that we can compute $\ve{m}_i$ and $\ve{m}_j$ by rotating the relative translation direction $\ve{c}_{ij}$ between $\ve{c}_i$ and $\ve{c}_j$, \ie $\ve{m}_i = \ve{R}(\theta_i)\ve{c}_{ij}$ and $\ve{m}_j = - \ve{R}(\theta_j)\ve{c}_{ij}$. Then \equref{equ:twoViewT_a} becomes}


\vspace{-0.02in}
\begin{equation} \label{equ:twoViewT_b} \small
\vspace{-0.02in}
\ve{p} = \frac{1}{2}\left( \ve{c}_i + s_{i}\ve{R}(\theta_i)\frac{\ve{c}_j-\ve{c}_i}{\left\|\ve{c}_j-\ve{c}_i\right\|}
   + \ve{c}_j + s_{j}\ve{R}(\theta_j)\frac{\ve{c}_i-\ve{c}_j}{\left\|\ve{c}_i-\ve{c}_j\right\|}\right).
\end{equation}
The two 3D rotation matrices $\ve{R}(\theta_i)$ and $\ve{R}(\theta_j)$ rotate the relative translation direction $\ve{c}_{ij}$ to the directions $\ve{m}_i$ and $\ve{m}_j$. Both rotations can be computed easily in the local pairwise reconstruction.
In addition, the two ratios $s_{i}/\left\|\ve{c}_j-\ve{c}_i\right\|$ and $s_{j}/\left\|\ve{c}_j-\ve{c}_i\right\|$
can be computed by the middle-point algorithm \cite{hartley2003book}.
Specifically, \cui{assuming unit baseline length, in the local coordinate system attached to one \cui{of} the cameras, $\ve{c}_i, \ve{c}_j$, $\ve{m}_i$, and $\ve{m}_j$ are all known.} Thus, we can solve $s_i$ and $s_j$ (they are actually the two ratios in \equref{equ:twoViewT_b} for general baseline length) from
\vspace{-0.02in}
\begin{equation}
\vspace{-0.02in}
(\ve{c}_i + s_i \ve{m}_i - \ve{c}_j - s_j \ve{m}_j) \times (\ve{m}_i \times \ve{m}_j) = 0.
\end{equation}
Here, $\times$ is the cross product of vectors.
Thus, \equref{equ:twoViewT_b} becomes,
\vspace{-0.02in}
\begin{equation} \label{equ:twoView_c}
\vspace{-0.02in}
\ve{p} = \frac{1}{2}\left( (\ve{A}_j^{ij} - \ve{A}_i^{ij})(\ve{c}_i - \ve{c}_j) +  \ve{c}_i + \ve{c}_j \right)
\end{equation}
where $\ve{A}_{i}^{ij} = s_{i}/||\ve{c}_j - \ve{c}_i|| \ve{R}(\theta_i)$ and
$\ve{A}_{j}^{ij}=s_j / ||\ve{c}_i - \ve{c}_j|| \ve{R}(\theta_j)$ are known matrices. This equation provides a linear constraint among positions of two camera centers and a scene point. Note this linear constraint minimizes a geometric error, the point-to-ray distance.



\subsection{Constraints from a feature track} \label{sec:feature_constraint}
If the same scene point $\ve{p}$ is visible in two image pairs $\ve{c}_i, \ve{c}_j$ and $\ve{c}_k, \ve{c}_l$ as shown in \figref{fig:triangulation} (b), we obtain two linear equations about $\ve{p}$'s position according to \equref{equ:twoView_c}. We can eliminate $\ve{p}$ from these equations to obtain a linear constraint among four camera centers as the following,
\vspace{-0.02in}
\begin{equation} \label{equ:MultipleView_C}
\vspace{-0.02in}
 (\ve{A}_j^{ij} - \ve{A}_i^{ij}) (\ve{c}_i - \ve{c}_j) + \ve{c}_ i + \ve{c}_j = (\ve{A}_l^{kl} - \ve{A}_k^{kl}) (\ve{c}_k - \ve{c}_l)  +  \ve{c}_k + \ve{c}_l.
\end{equation}
Given a set of images, we build feature tracks and collect such linear equations from camera pairs on the same feature track. Solving these equations will provide a linear global solution of camera positions. To resolve the gauge ambiguity, we set the orthocenter of all cameras at origin when solving these equations.

\equref{equ:MultipleView_C} elegantly correlates the scales of translation (\ie baseline length\cui{$\left\|\ve{c}_i-\ve{c}_j\right\|$ and $\left\|\ve{c}_k-\ve{c}_l\right\|$}) for camera pairs sharing a common scene point.
For example, in the \emph{Seville} data in \figref{fig:motivation} (bottom row),  $\ve{c}_i, \ve{c}_j$ could come from one popular viewpoint of the building, and $\ve{c}_k, \ve{c}_l$ could come from a different viewpoint. As long as there is a feature track linking them together, \equref{equ:MultipleView_C} provides constraints among the baseline lengths of these far apart cameras.
In comparison, triplet-based methods (\eg \cite{Sim2006,jiang2013g,moulon2013gf}) can only propagate the scale information over camera triplets sharing an edge. 
Clearly, the scale information can propagate further along feature tracks. 
Therefore, this new formulation can reconstruct images with \todo{weak association better} than triplet-based methods.

\subsection{Feature tracks selection}  \label{sec:feature_track_selection}
Since there are usually abundant feature tracks to solve camera positions, we carefully choose the most reliable ones to enhance system robustness.
For better feature matching quality, we only consider feature correspondences that are inliers of essential matrix fitting. We sort all feature tracks by their lengths in descending order, and then try to find a small set of tracks that could cover all connected cameras at least $K$ times. (We set $K = 30$ in our experiments.) Please see our supplementary material for the pseudo-code of the feature tracks selection.

For a feature track with $N_t$ cameras, there are usually more than $N_t-1$ EGs on it. So we select the most reliable ones to construct equations. We consider the match graph formed by these cameras, where two cameras are connected when their essential matrix is known. Since we only consider feature correspondences passing essential matrix verification, this graph has only one connected component. We weight each graph edge by $\frac{1}{M}+\alpha\frac{1}{\theta}$,  where $M$ is the number of feature matches between two images, and $\theta$ is the triangulation angle. The combination weight $\alpha$ is fixed at $0.1$ in our experiments. We take the minimum spanning tree of this graph, and randomly choose two edges from the tree to build a linear equation until each edge is used twice.

\section{Robust Estimation by $L_1$ Norm} \label{sec:l1}

Our linear global method requires solving a linear system like $\ve{A}\ve{x} = 0$ to estimate camera centers. 
\cuif{$\ve{x}$ represents an unknown vector formed by concatenating all camera positions, and  $\ve{A}$ is the coefficient matrix formed by collecting \equref{equ:MultipleView_C} from feature tracks. }

The 3D reconstruction process is noisy and involves many outliers, both in  essential matrices and feature correspondences. 
We enhance  system robustness by minimizing the $L_1$ norm, instead of the conventional $L_2$ norm. In other words, we solve the following optimization problem,
\vspace{-0.02in}
\begin{equation} \label{equ:L1}
\vspace{-0.02in}
\arg\min_{\ve{x}}\left\| \ve{A}\ve{x}\right\|_1, \qquad s.t. \quad \ve{x}^\top \ve{x}=1.
\end{equation}
This problem might be solved by iterative reweighted total least squares, which is often slow and requires good initialization. \cuif{Recently, Ferraz \etal \cite{ferraz2014} proposed a robust method to discard outliers, while it is not applicable to our large sparse homogeneous system.}  
We capitalize on the recent alternating direction method of multipliers (ADMM) \cite{Boyd2011ADMM} for better efficiency and large convergence region. Due to the quadratic constraint, \ie $\ve{x}^\top \ve{x} = 1$, the original ADMM algorithm cannot be directly applied to our problem. We linearize the optimization problem in the inner loop of ADMM to solve \equref{equ:L1}.


Let $\ve{e} = \ve{A}\ve{x}$, the augmented Lagrangian function of \equref{equ:L1} is
\vspace{-0.05in}
\begin{equation} \label{equ:L}
\vspace{-0.05in}
\begin{aligned}
L(\ve{e},\ve{x},\lambda) = \left\| \ve{e}\right\|_1 + \left\langle \lambda, \ve{A} \ve{x}-\ve{e} \right\rangle + \frac{\beta}{2}\left\| \ve{A}\ve{x}-\ve{e}\right\|^2 , ~~~~~~~s.t. \qquad \ve{x}^\top \ve{x}=1,
\end{aligned}
\end{equation}
where $\lambda$ is the Lagrange multiplier, $\left\langle \cdot,\cdot \right\rangle$ is the inner product, and $\beta>0$ is a parameter controlling the relaxation.

We then iteratively optimize $\ve{e}$, $\ve{x}$, and $\lambda$ in \equref{equ:L}. In each iteration, we update $\ve{e}_{k+1}$, $\ve{x}_{k+1}$, $\lambda_{k+1}$ according to the following scheme,
\vspace{-0.05in}
\begin{eqnarray}
\small
\ve{e}_{k+1} &=& \arg \min_{\ve{e}} L(\ve{e}, \ve{x}_k, \lambda_k)  =  \arg \min_{\ve{e}} \left\| \ve{e}\right\|_1 + \left\langle \lambda_{k}, \ve{A} \ve{x}_k - \ve{e} \right\rangle + \frac{\beta}{2}\left\| \ve{A} \ve{x}_k - \ve{e}\right\|^2, \label{equ:update_e} \\
\ve{x}_{k+1} &=&  \arg \min_{\ve{x} \in \Omega} L(\ve{e}_{k+1}, \ve{x}, \lambda_k)  =  \arg \min_{\ve{x} \in \Omega} \left\langle \lambda_k, \ve{A} \ve{x}- \ve{e}_{k+1} \right\rangle + \frac{\beta}{2}\left\| \ve{A} \ve{x} - \ve{e}_{k+1}\right\|^2, \label{equ:update_x} \\ [0.2cm]
\lambda_{k+1} &=& \lambda_k + \beta(\ve{A} \ve{x}_{k+1} - \ve{e}_{k+1}),  \label{equ:update_lambda}
\end{eqnarray}
where $\Omega:=\left\lbrace \ve{x}^\top \ve{x} = 1| \ve{x}\in \mathbb{R}^n\right\rbrace$. A closed-form solution \cite{lin2010augmented} exists for the minimization of  \equref{equ:update_e}. \cuif{(Please see Appendix A for the formula.)} 
Solving \equref{equ:update_x} is hard because of the quadratic constraint on $\ve{x}$. 
Therefore, we linearize \equref{equ:update_x} and derive a closed-form solution as,
\vspace{-0.025in}
\begin{equation} \label{equ:update_x_linear_sol}
\vspace{-0.025in}
\ve{x}_{k+1} = C / \left\| C\right\|^2 ,
\end{equation}
where $C = \ve{x}_k - \frac{1}{\eta} \ve{A}^\top ( \ve{A} \ve{x}_k - \ve{e}_{k+1})-\frac{1}{\beta\eta} \ve{A}^\top \lambda^k$, and $\eta > \sigma_{max}(\ve{A}^\top\ve{A})$. \cuif{(Please see Appendix B for more details.)}
\cui{In order to speed up convergence \cite{lin2011linearized}, we adopt a dynamic parameter $\beta$ as,}
\vspace{-0.025in}
\begin{equation} \label{equ:update_beta}
\vspace{-0.025in}
\beta_{k+1} = min{\left\lbrace \beta_{max},\rho\beta_k\right\rbrace },
\end{equation}
where $\rho >1$. We set $\rho$ as \todo{$1.01$ or $1.1$ for sequential data and Internet data respectively} in our experiments. 
\alref{alg:l1} summarizes our linearized ADMM algorithm.

\vspace{-0.1in}
\begin{algorithm}
\caption{Our linearized ADMM for \equref{equ:L1}.}
	\label{alg:l1}
\begin{algorithmic}[1]
\State \textbf{Initialize}: Set $\ve{x}_0$ as to the $L_2$ solution (\ie the eigenvector with smallest eigenvalue of $\ve{A}$), $\ve{e}_0 = \mathbf{0}$, $\lambda_0 = \mathbf{0}$, $\beta_0 = 10^{-6}$;
\While {not converged,}
\State \textbf{Step 1}: Update $\ve{e}$ by solving \equref{equ:update_e};
\State \textbf{Step 2}: Update $\ve{x}$ by solving \equref{equ:update_x_linear_sol};
\State \textbf{Step 3}: Update $\lambda$ by solving \equref{equ:update_lambda};
\State \textbf{Step 4}: Update $\beta$ by solving \equref{equ:update_beta};
\EndWhile
\end{algorithmic}
\end{algorithm}

\vspace{-0.1in}
\section{Experiments} \label{sec:experiment}

\begin{table} [t]\centering \footnotesize
\begin{tabular} 
{|>{\raggedright\arraybackslash}m{0.135\linewidth}
|>{\centering\arraybackslash}m{0.08\linewidth}
|>{\centering\arraybackslash}m{0.067\linewidth}
|>{\centering\arraybackslash}m{0.091\linewidth}
|>{\centering\arraybackslash}m{0.05\linewidth}
|>{\centering\arraybackslash}m{0.12\linewidth}
|>{\centering\arraybackslash}m{0.089\linewidth}
|>{\centering\arraybackslash}m{0.03\linewidth}
|>{\centering\arraybackslash}m{0.03\linewidth}
|}
\cline{1-9}
\multirow{2}{*}{Dataset} & \multicolumn{8}{c|}{$c_{err}$ (GT,mm)} \\ \cline{2-9}

  & Rother\cite{Rother2003} & Jiang\cite{jiang2013g} & Moulon\cite{moulon2013gf}
          & Arie\cite{Arie2012} & VisualSFM\cite{wu2013visualsfm} & 1DSfM\cite{Wilson2014} & $L_2$ & $L_1$ \\ \hline

\textit{fountain-P11}  & \textbf{2.5} & 3.1 & \textbf{2.5} & 2.9 &3.6 & 33.5 & \textbf{2.5}& \textbf{2.5}\\ \hline
\textit{Herz-Jesu-P25} & \textbf{5.0} & 7.5 & 5.3 & 5.3 & 5.7  & 36.3 & \textbf{5.0} & \textbf{5.0}\\ \hline
\textit{castle-P30}    & 347.0 & 72.4 & 21.9 & - & 70.7 & - & 21.6 & \textbf{21.2} \\ \hline
\end{tabular}
\vspace{-0.13in}
\caption{Reconstruction accuracy comparison on benchmark data with ground truth (GT) camera intrinsics.}
\vspace{-0.1in}
\label{tab:benchmark1}
\end{table}

\begin{table}[t]\centering \footnotesize
\begin{tabular} {|l|c|c|c|c|c|c|c|}
\cline{1-8}
\multirow{2}{*}{Dataset} & \multicolumn{7}{c|}{$c_{err}$ (EXIF,mm)} \\ \cline{2-8}
    & Rother\cite{Rother2003} & Jiang\cite{jiang2013g} & Arie\cite{Arie2012} & VisualSFM\cite{wu2013visualsfm} & 1DSfM\cite{Wilson2014} & $L_2$ & $L_1$\\ \hline
\textit{fountain-P11}   & 23.3 & 14.0  & 22.6 & 20.7 & 32.2 & \textbf{6.9} & 7.0  \\ \hline
\textit{Herz-Jesu-P25}  & 49.5 & 64.0  & 47.9 & 45.3 & 64.9 & \textbf{25.5} & 26.2  \\ \hline
\textit{castle-P30}   & 2651.8 & 235.0 & - & 190.1 & - & 317.3 &  \textbf{166.7}  \\
\hline
\end{tabular}
\vspace{0.03in}
\caption{Reconstruction accuracy comparison on benchmark data with approximate intrinsics from EXIF. The results by Moulon\cite{moulon2013gf} are not available. }
\label{tab:benchmark2}
\end{table}

\subsection{Evaluation on benchmark data} \label{sec:benchmark}

We compare our method with  VisualSFM \cite{wu2013visualsfm}, and several global SfM methods on the benchmark data provided in \cite{Strecha2008}. We use both ground truth camera intrinsics and approximate intrinsics from EXIF in the experiment. We implement the method in \cite{Rother2003} by ourselves. The results on VisualSFM \cite{wu2013visualsfm} and 1DSfM \cite{Wilson2014} are obtained by running the codes provided by the authors.  To evaluate the $L_1$ norm optimization, we also experiment the conventional $L_2$ norm optimization instead of the $L_1$ norm in \equref{equ:L1}. The results are indicated as $L_1$ and $L_2$ respectively.

We summarize all the results in \tabref{tab:benchmark1} and \tabref{tab:benchmark2}. All results are evaluated after the final bundle adjustment. 
Our method generally produces the smallest errors with either ground truth intrinsics or approximate ones from EXIF.
The $L_2$ and $L_1$ methods produce similar results on the  \textit{fountain-P11} and \textit{Herz-Jesu-P25} data, since these data have few outliers.
But the $L_1$ method outperforms $L_2$ significantly on the \textit{castle-P30} data, whose essential matrix estimation suffers from repetitive scene structures. \cui{The noisy epipolar geometries also cause bad performance of the method in \cite{Rother2003} on the \textit{castle-P30} data, which solves cameras and scene points together by minimizing an algebraic error. In comparison, our method minimizes a geometric error, \todo{which is the point-to-ray distance}, and achieves better robustness.}

\subsection{Experiment on sequential data}

\begin{figure} \centering
\begin{tabular}{cccc}

\multicolumn{4}{c}{\includegraphics[width=0.85\linewidth]{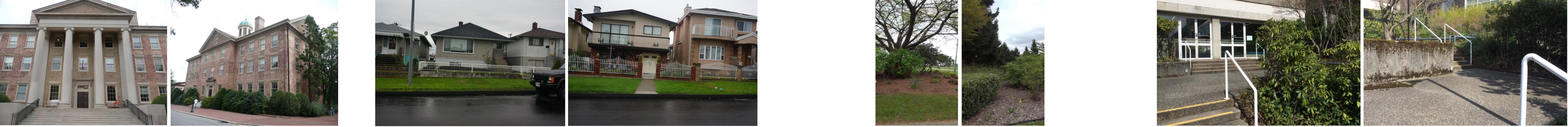}}\\[-0.05cm]
	
 \includegraphics[width=0.18\linewidth]{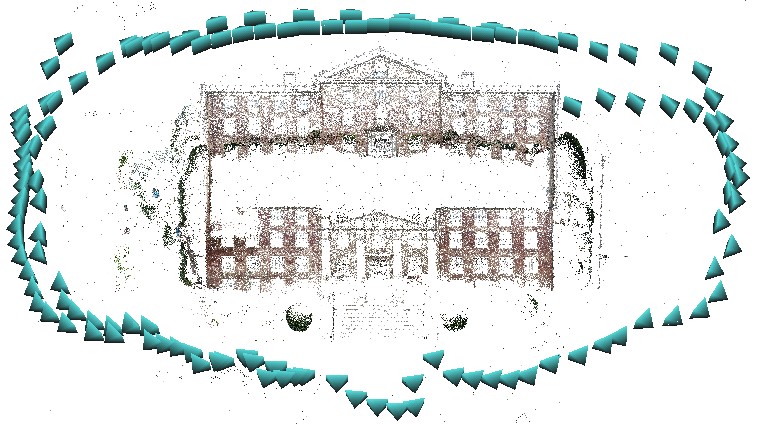} & 
 \includegraphics[width=0.19\linewidth]{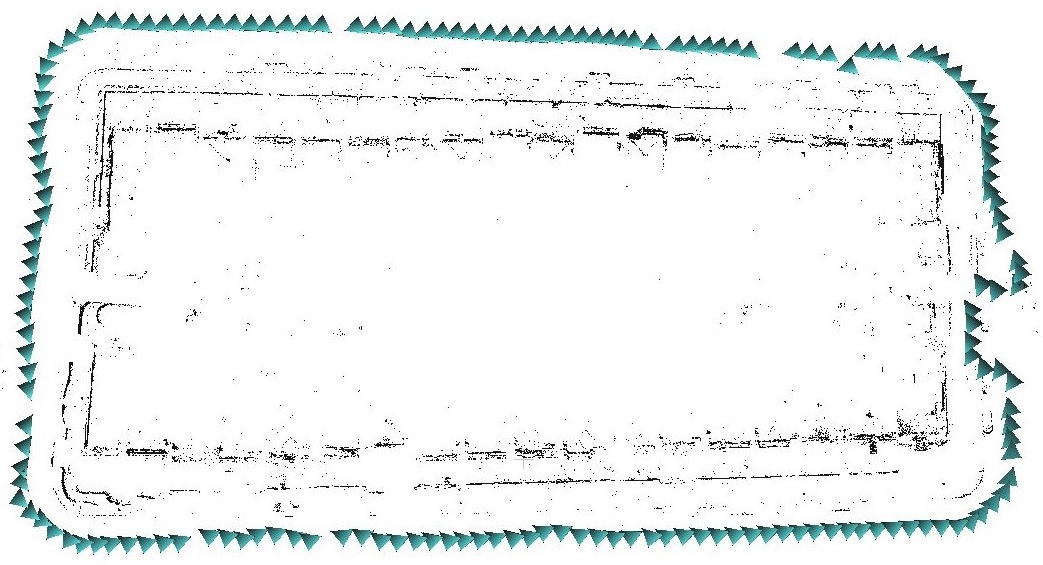} & 
 \includegraphics[width=0.18\linewidth]{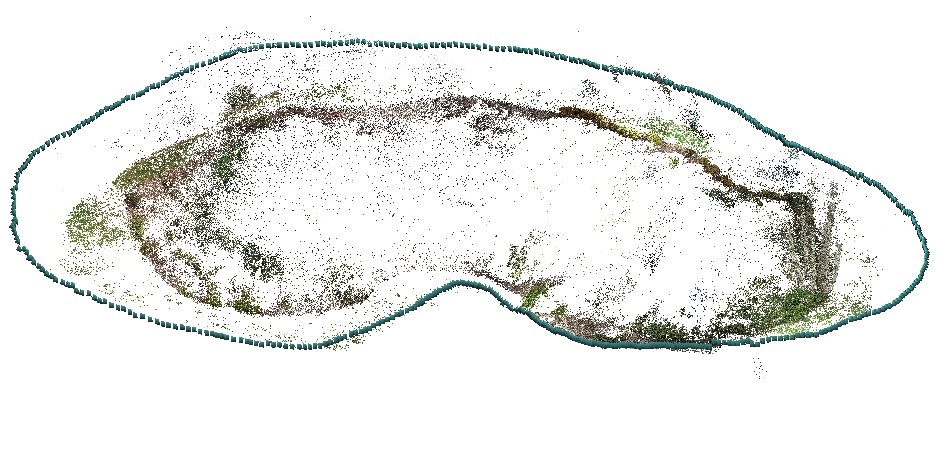}& 
 \includegraphics[width=0.19\linewidth]{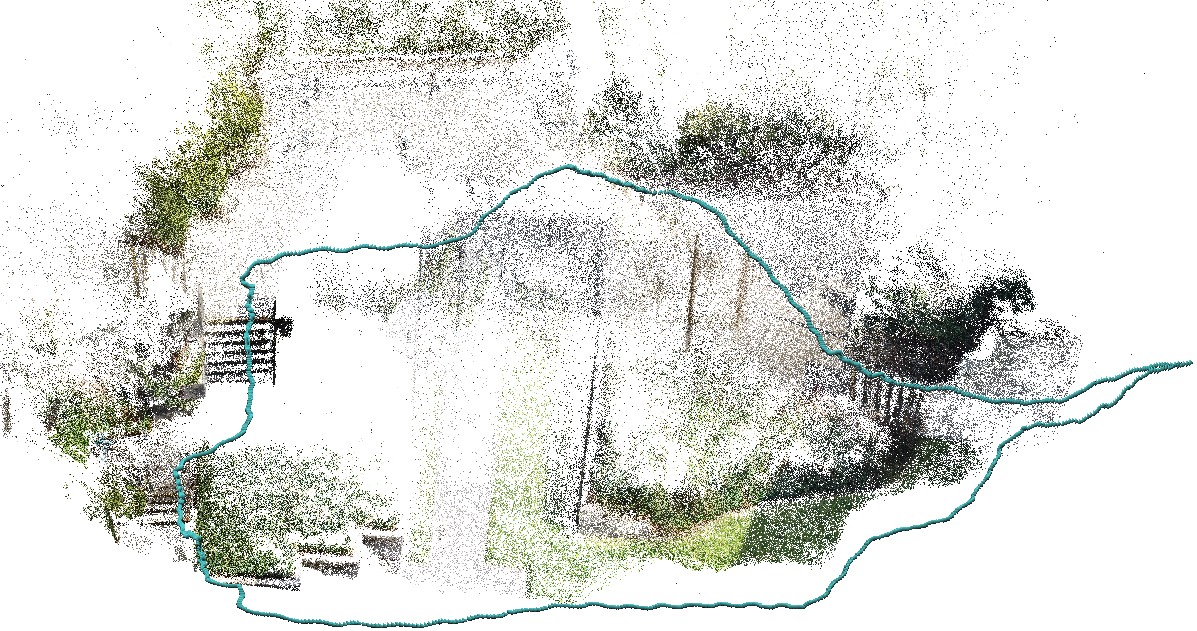}\\ [-0.05cm]

 \includegraphics[width=0.18\linewidth]{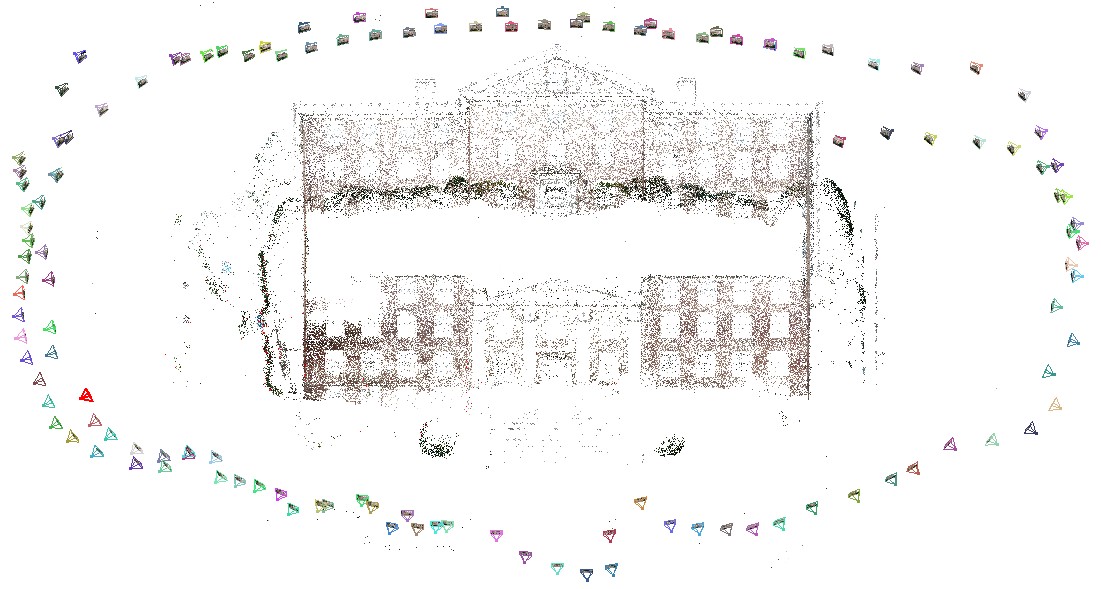}&
 \includegraphics[width=0.19\linewidth]{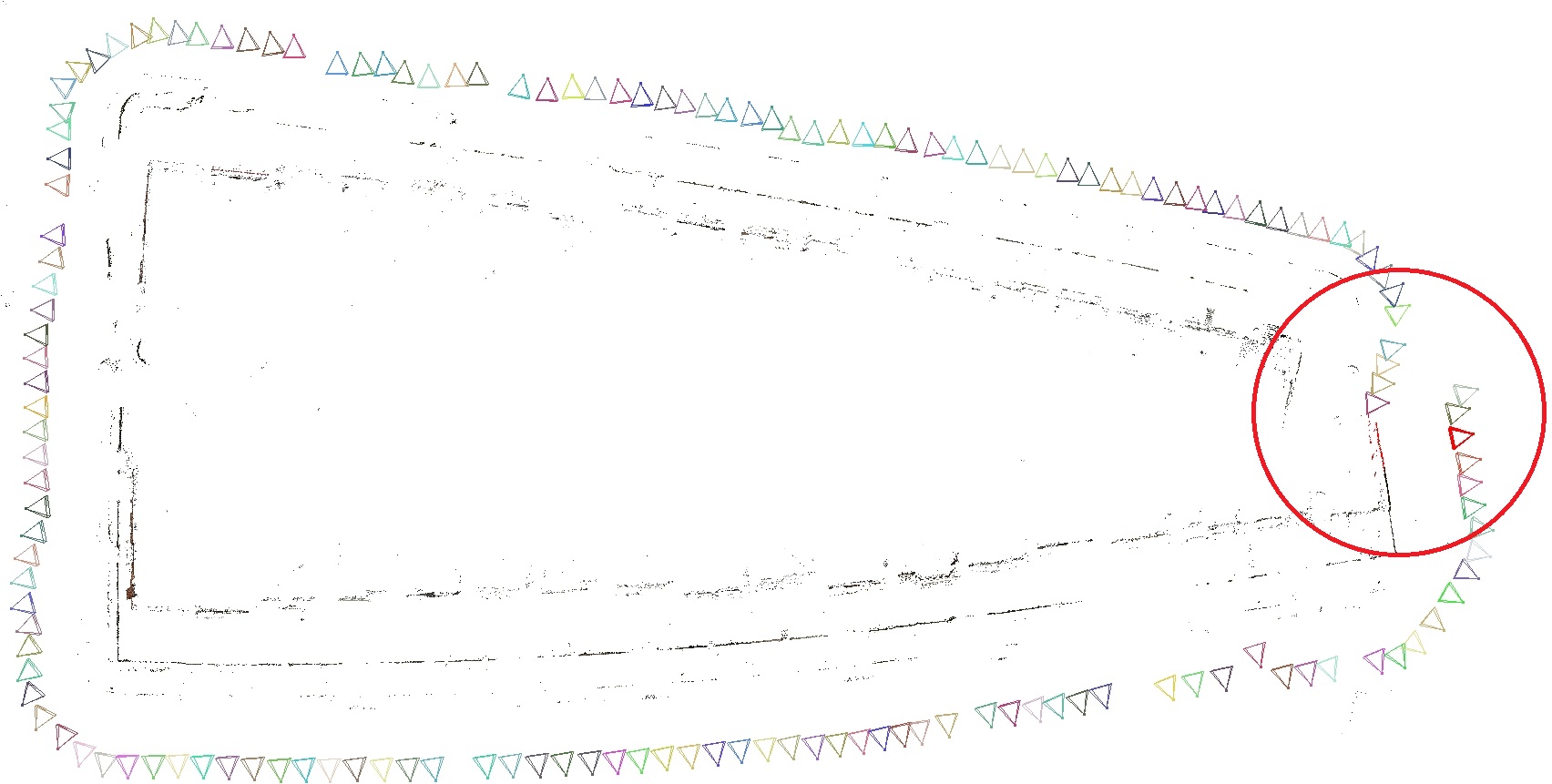}&
 \includegraphics[width=0.18\linewidth]{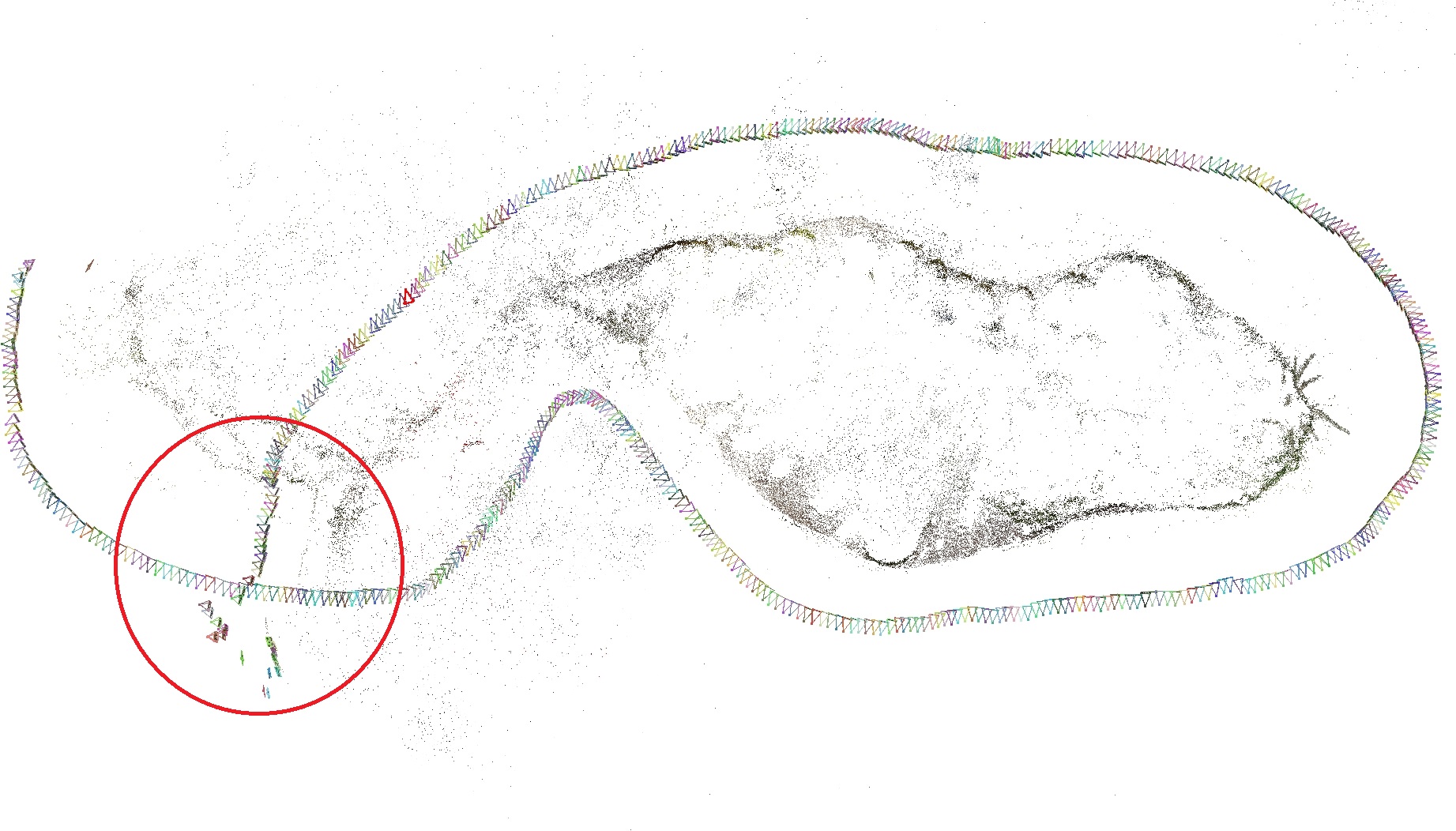}&
 \includegraphics[width=0.19\linewidth]{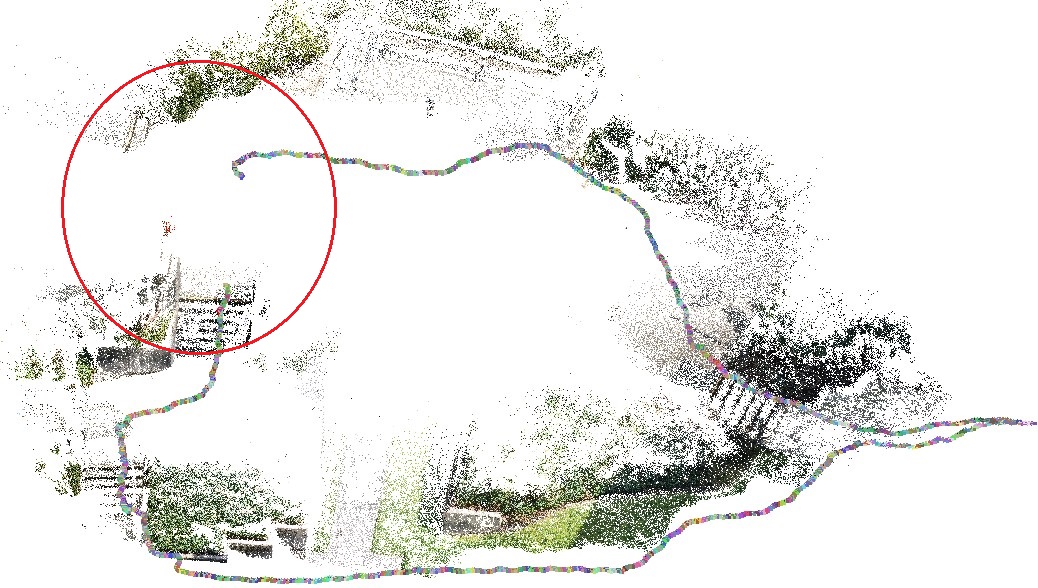}\\[-0.05cm]

 \includegraphics[width=0.18\linewidth]{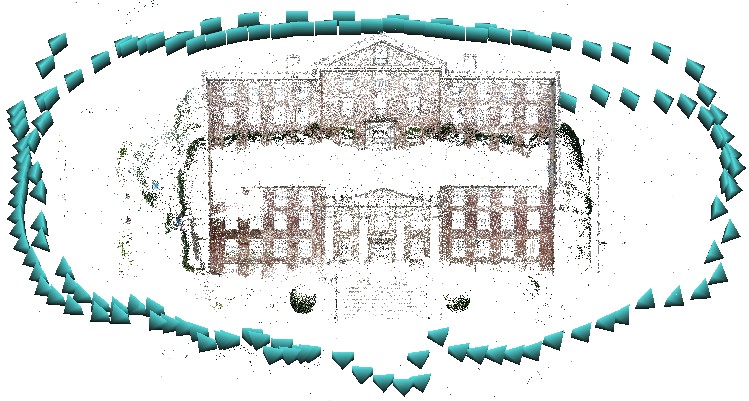}&
 \includegraphics[width=0.19\linewidth]{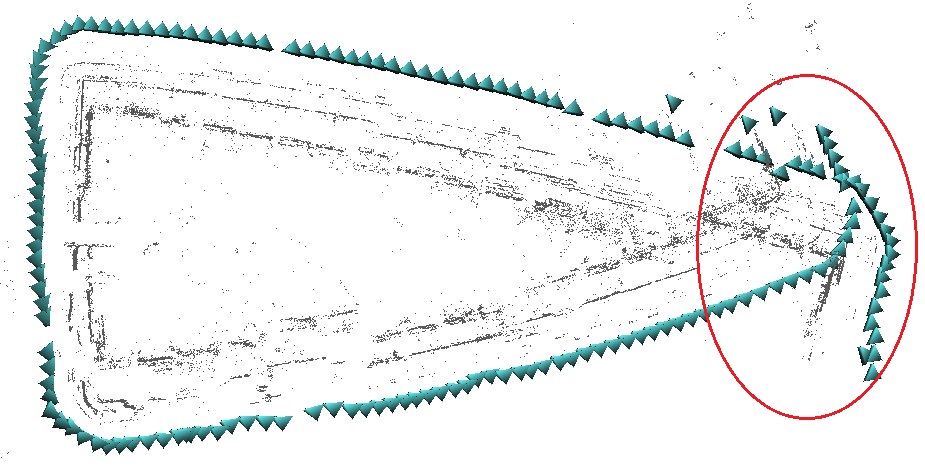}&
 \includegraphics[width=0.18\linewidth]{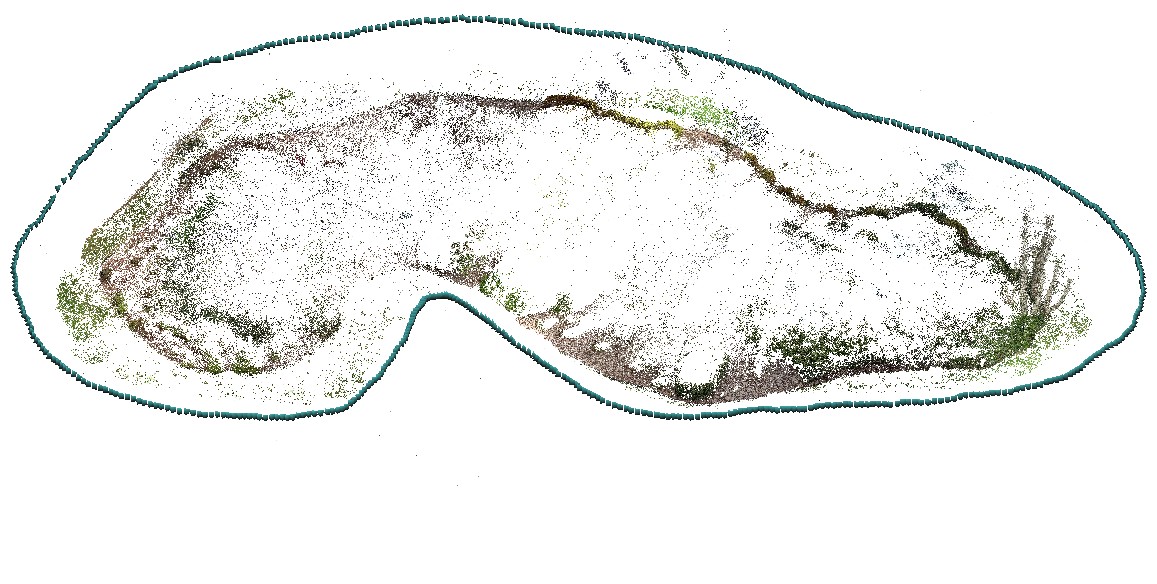}&
 \includegraphics[width=0.19\linewidth]{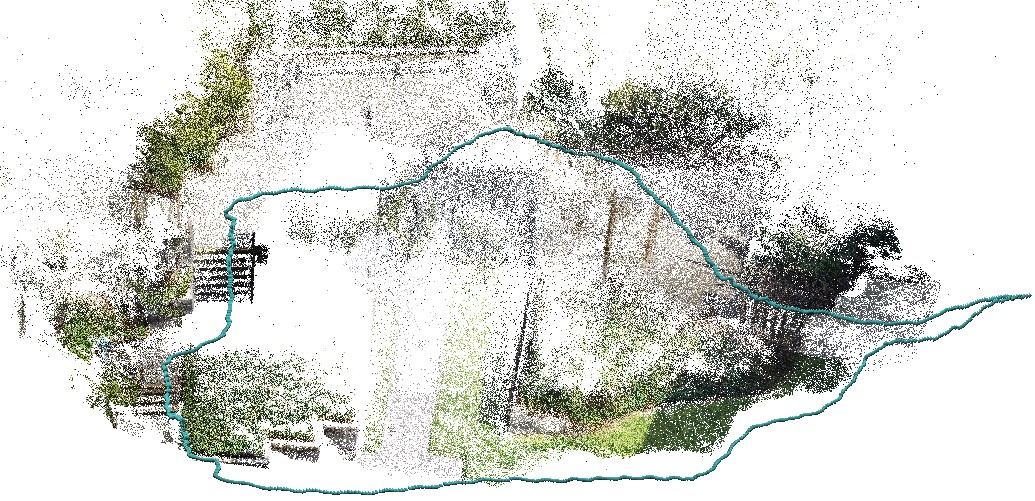}\\


 (a) \emph{Building} &  (b) \emph{Street} & (c) \emph{Park} & (d) \emph{Stair}\\
\end{tabular}
\caption{Evaluation on sequential data. From top to bottom, each row shows sample input images, 3D reconstructions generated by our method, VisualSFM \cite{wu2013visualsfm}, and the least unsquared deviations (LUD) method \cite{Ozyesil2015} respectively.}
\label{fig:SequentialResults} 
\end{figure}

\figref{fig:SequentialResults} summarizes our experiment on sequentially captured data and compares our method with an incremental method, VisualSFM \cite{wu2013visualsfm}, and some recent global methods \cite{Wilson2014, Ozyesil2015}. The test data \emph{Building}, \emph{Street}, \emph{Park} and \emph{Stair} have 128, 168, 507 and 1221 input images respectively. Generally speaking, VisualSFM \cite{wu2013visualsfm} suffers from large drifting errors when the input essential matrices are noisy or when the image sequence is long, as shown in the third row of \figref{fig:SequentialResults}. The drifting errors in the \emph{Park} and \emph{Stair} examples are severe because of the poor essential matrix estimation due to poor feature localization in their tree images. We do not include results from 1DSfM \cite{Wilson2014} in \figref{fig:SequentialResults}, since it fails on all these sequential data. Its result on the \emph{Street} data is shown in \figref{fig:motivation} (left of top row). 1DSfM cannot handle these examples because it is designed for Internet images which tend to have $O(n^2)$ essential matrices for $n$ images \footnote{This is according to our discussion with the authors of 1DSfM \cite{Wilson2014}.}. The least unsquared deviations (LUD) method \cite{Ozyesil2015} generates distortion on the \emph{Street} example, because it degenerates at collinear motion. In comparison, our method does not have visible drifting and is robust to collinear motion and weak image association.

\begin{table*}\centering \scriptsize  
\begin{tabular} 
{|>{\raggedright\arraybackslash}m{0.138\linewidth}
|>{\centering\arraybackslash}m{0.021\linewidth}
|>{\centering\arraybackslash}m{0.021\linewidth}
|>{\centering\arraybackslash}m{0.021\linewidth}
|>{\centering\arraybackslash}m{0.021\linewidth}
|>{\centering\arraybackslash}m{0.021\linewidth}
|>{\centering\arraybackslash}m{0.021\linewidth}
|>{\centering\arraybackslash}m{0.021\linewidth}
|>{\centering\arraybackslash}m{0.021\linewidth}
|>{\centering\arraybackslash}m{0.021\linewidth}
|>{\centering\arraybackslash}m{0.021\linewidth}
|>{\centering\arraybackslash}m{0.021\linewidth}
|>{\centering\arraybackslash}m{0.021\linewidth}
|>{\centering\arraybackslash}m{0.021\linewidth}
|>{\centering\arraybackslash}m{0.021\linewidth}
|>{\centering\arraybackslash}m{0.021\linewidth}
|}
\hline
\multicolumn{2}{|c|}{Dataset}
& \multicolumn{4}{c|}{1DSfM\cite{Wilson2014}}
& \multicolumn{4}{c|}{LUD\cite{Ozyesil2015}}
& \multicolumn{2}{c|}{$L_2$}
& \multicolumn{4}{c|}{$L_1$}
\\ \hline

Name & $N_i$ &$N_{c}$ & $\tilde{x}$ & $\tilde{x}_{BA}$& $\bar{x}_{BA}$ & $N_{c}$ & $\tilde{x}$ & $\tilde{x}_{BA}$& $\bar{x}_{BA}$ & $N_{c}$ & $\tilde{x}$   & $N_{c}$ & $\tilde{x}$ & $\tilde{x}_{BA}$& $\bar{x}_{BA}$    \\ \hline

\emph{Alamo} &613			&	529	&	1.1	&			\best{0.3}	&	2e7	&	547	&	\best{0.4}	&					\best{0.3}	&	\best{2.0}	&			496	&	0.5	&		500	&	0.6	&	0.5	&	3.7	\\ \hline
\emph{Ellis Island} &242		&	214	&	3.7	&			\best{0.3}	&	3.0	&	-	&	-	&						- &		- &			183	&	9.4	&		211	&	\best{3.1}	&	0.6	&	\best{1.8}	\\ \hline
\emph{Montreal N.D.} &467	&	427	&	2.5	&			\best{0.4}	&	\best{1.0}	&	435	&	\best{0.5}	&					\best{0.4}	&	\best{1.0}	&			424	&	0.8	&		426	&	0.8	&	\best{0.4}	&	1.1	\\ \hline
\emph{Notre Dame} &552		&	507	&	10	&			1.9	&	7.0	&	536	&	\best{0.3}	&					\best{0.2}	&	\best{0.7}	&			537	&	\best{0.3}	&		539	&	\best{0.3}	&	\best{0.2}	&	0.8	\\ \hline
\emph{NYC Library} &369		&	295	&	2.5	&			\best{0.4}	&	\best{1.0}	&	320	&	2.0	&					1.4	&	7.0	&			-	&	-	&		288	&	\best{1.4}	&	0.9	&	6.9	\\ \hline
\emph{Piazza del Popolo} &350	&	308	&	3.1	&			2.2	&	2e2	&	305	&	\best{1.5}	&	\best{1.0}	&	4.0	&			302	&	3.6	&		294	&	2.6	&	2.4	&	\best{3.2}	\\ \hline
\emph{Tower of London} &499	&	414	&	11	&			\best{1.0}	&	40	&	425	&	4.7	&					3.3	&	10	&			311	&	17	&		393	&	\best{4.4}	&	1.1	&	\best{6.2}	\\ \hline
\emph{Vienna Cathedral} &897	&	770	&	6.6	&			\best{0.4}	&	2e4	&	750	&	5.4	&					4.4	&	10	&			574	&	3.6	&		578	&	\best{3.5}	&	2.6	&	\best{4.0}	\\ \hline
\emph{Yorkminster}	&450	&	401	&	3.4	&			\best{0.1}	&	5e2	&	404	&	\best{2.7}	&				1.3	&	\best{4.0}	&			333	&	3.9	&		341	&	3.7	&	3.8	&	14	\\ \hline																																						

\end{tabular}
\vspace{0.02in}
\caption{Comparison with \cite{Wilson2014} on challenging data.  $N_{i}$ denotes the number of cameras in the largest connected component of our EG graph, and $N_{c}$ denotes the number of reconstructed cameras. $\tilde{x}$ denotes the median error before BA. $\tilde{x}_{BA}$ and $\bar{x}_{BA}$ denote the median error and the average error after BA respectively. The errors are the distances in meters to corresponding cameras computed by an incremental SfM method \cite{snavely2006}.}
\label{tab:compare_1dsfm}

\end{table*}

\begin{table*}\centering \scriptsize 
\begin{tabular} {|l|c|c|c|c|c|c|c|} \hline
\multirow{2}{*}{\parbox{2cm}{Dataset}}
& \multicolumn{2}{c|}{1DSfM\cite{Wilson2014}}
& \multicolumn{2}{c|}{LUD\cite{Ozyesil2015}}
& \multicolumn{1}{c|}{Bundler\cite{snavely2006}}
& \multicolumn{2}{c|}{$L_1$}

 \\ \cline{2-8}

& $T_{BA}$ & $T_{\Sigma}$  & $T_{BA}$ & $T_{\Sigma}$ & $T_{\Sigma}$ & $T_{BA}$ & $T_{\Sigma}$  
\\ \hline
\emph{Alamo}			&	752	&	910	&	133	&	750	&	1654	&	362	&	621	\\ \hline
\emph{Ellis Island}			&	139	&	171	&	-	&	-	&	1191	&	64	&	95	\\ \hline
\emph{Montreal N.D.}			&	1135	&	1249	&	167	&	553	&	2710	&	226	&	351	\\ \hline
\emph{Notre Dame}			&	1445	&	1599	&	126	&	1047	&	6154	&	793	&	1159	\\ \hline
\emph{NYC Library}			&	392	&	468	&	54	&	200	&	3807	&	48	&	90	\\ \hline
\emph{Piazza del Popolo}			&	191	&	249	&	31	&	162	&	1287	&	93	&	144	\\ \hline
\emph{Tower of London}			&	606	&	648	&	86	&	228	&	1900	&	121	&	221	\\ \hline
\emph{Vienna Cathedral}			&	2837	&	3139	&	208	&	1467	&	10276	&	717	&	959	\\ \hline
\emph{Yorkminster}			&	777	&	899	&	148	&	297	&	3225	&	63	&	108	\\ \hline

\end{tabular}
\vspace{0.02in}
\caption{Running times in seconds for the Internet data. $T_{BA}$ and $T_{\Sigma}$ denote the final bundle adjustment time and total running time respectively.}
\label{tab:compare_1dsfm_time}
\end{table*}

\subsection{Experiment on unordered Internet data}
The input epipolar geometries for Internet data are quite noisy because of the poor feature matching. Besides using $L_1$ optimization, we adopt two additional steps to improve the robustness of our method. After solving camera orientations by the method in \cite{chatterjee2013efficient}, we further filter input essential matrices with the computed camera orientations. Specifically, for each camera pair, we compare their relative rotation from their global orientation with the relative rotation encoded in their essential matrix. If the difference is larger than a threshold $\varphi_2$ (\cui{set to $5$ or $10$} degrees), we discard that essential matrix.  What's more, we refine the relative translations with the camera orientations fixed using the method in \cite{kneip2011robust}.

We test our method on the challenging Internet data released by \cite{Wilson2014} and compare our method with several global methods. We use the results of an optimized incremental SfM system based on Bundler \cite{snavely2006} as the reference `ground-truth' and compute the camera position errors for evaluation. 
As shown in \tabref{tab:compare_1dsfm}, our method with $L_1$ optimization has smaller initial median errors than \cite{Wilson2014} and comparable errors with \cite{Ozyesil2015}. Our method with $L_1$ optimization performs better than $L_2$ solutions, which shows the effectiveness of the proposed $L_1$ method. All the methods have similar results after the final bundle adjustment. 

\tabref{tab:compare_1dsfm_time} lists the running time of different methods. All our experiments were run on a machine with two 2.4GHz Intel Xeon E5645 processors with 16 threads enabled. We cite the running time for \cite{Wilson2014}, \cite{Ozyesil2015} and \cite{snavely2006} for comparison. Our method is around 10 times faster than the optimized incremental method  \cite{snavely2006} and also faster than the global methods \cite{Wilson2014, Ozyesil2015}.



\section{Conclusion} \label{sec:conclusion}

We derive a novel linear method for global camera translation estimation. This method is based on a novel
position constraint on cameras linked by a feature track, which minimizes a geometric error and propagates the scale information across far apart camera pairs. In this way, our method works well even on weakly associated images. The final linear formulation does not involve coordinates of scene points, so it is easily scalable and computationally efficient. We further develop an $L_1$ optimization method to make the solution robust to outlier essential matrices and feature correspondences. Experiments on various data and comparison with recent works demonstrate the effectiveness of this new algorithm.

\section*{}
\vspace{-0.58in}
\textbf{Acknowledgements.} This work is supported by the NSERC Discovery grant 611664, Discovery Acceleration Supplements 611663, and the HCCS research grant at the ADSC from Singapore's Agency for Science, Technology and Research (A*STAR).

\appendix

\section*{Appendix}

\paragraph{A. Solution for \equref{equ:update_e}.} \label{app:update_e_derivation}
From \equref{equ:update_e}, we have
\vspace{-0.05in}
\begin{eqnarray}
\vspace{-0.05in}
\small
\ve{e}_{k+1} & = &\arg \min_{\ve{e}}  \frac{1}{\beta} \left\| \ve{e}\right\|_1  + \frac{1}{2}\left\| \ve{A} \ve{x}_k - \ve{e} + \frac{\lambda_{k}}{\beta} \right\|^2 \nonumber \\
& = &\arg \min_{\ve{e}}  \epsilon \left\| \ve{e}\right\|_1  + \frac{1}{2}\left\| \ve{e} - \ve{u} \right\|^2,
 \label{equ:update_e_further}
\end{eqnarray}
where $ \epsilon = \frac{1}{\beta}$, and $\ve{u} = \ve{A} \ve{x}_k + \frac{\lambda_{k}}{\beta}$. According to \cite{lin2010augmented}, the solution for \equref{equ:update_e_further}  is
\vspace{-0.05in}
\begin{equation}
\vspace{-0.05in}
\small
\ve{e}_{k+1}^{i} = \begin{cases}
\ve{u}^{i} - \epsilon, & if~~ \ve{u}^{i} > \epsilon, \\
\ve{u}^{i} + \epsilon & if ~~ \ve{u}^{i} < -\epsilon,\\
0, & otherwise,
  \end{cases}
\end{equation}
where $\ve{e}_{k+1}^{i}$ and $\ve{u}^{i}$ are the $i$-th element of $\ve{e}$ and $\ve{u}$.

\paragraph{B. Derivation of \equref{equ:update_x_linear_sol}.}\label{app:linearization_derivation}
We linearize the quadratic term $\frac{\beta}{2}\left\| \ve{A}\ve{x} - \ve{e}_{k+1}\right\|^2$ in \equref{equ:update_x} at $\ve{x}_k$, which gives
\vspace{-0.05in}
\begin{equation}
\vspace{-0.05in}
\small
\begin{split} 
\ve{x}_{k+1}  &=  \arg \min_{\ve{x} \in \Omega} \left\langle \ve{A}^\top \lambda_k, \ve{x} \right\rangle + \left\langle \beta \ve{A}^\top(\ve{A}\ve{x}_k - \ve{e}_{k+1}), \ve{x}-\ve{x}_k \right\rangle +\frac{\beta \eta}{2}\left\| \ve{x}-\ve{x}_{k}\right\|^2\\
&= \arg \min_{\ve{x} \in \Omega} \frac{\beta \eta}{2} \left\| \ve{x}- C \right\|^2, 
 \end{split}
\end{equation}
where $\Omega:=\left\lbrace \ve{x}^\top\ve{x} = 1| \ve{x}\in \mathbb{R}^n\right\rbrace$, $C = \ve{x}_k - \frac{1}{\eta} \ve{A}^\top ( \ve{A} \ve{x}_k - \ve{e}_{k+1})-\frac{1}{\beta\eta} \ve{A}^\top \lambda^k$, and $\eta > \sigma_{max}(\ve{A}^\top\ve{A})$ is a proximal parameter. Therefore, we can get \equref{equ:update_x_linear_sol} directly.

\bibliography{egbib}

\begin{thebibliography}{48}
\providecommand{\natexlab}[1]{#1}
\providecommand{\url}[1]{\texttt{#1}}
\expandafter\ifx\csname urlstyle\endcsname\relax
  \providecommand{\doi}[1]{doi: #1}\else
  \providecommand{\doi}{doi: \begingroup \urlstyle{rm}\Url}\fi

\bibitem[Agarwal et~al.()Agarwal, Mierle, and Others]{ceres-solver}
S.~Agarwal, K.~Mierle, and Others.
\newblock Ceres solver.
\newblock \url{http://ceres-solver.org}.

\bibitem[Agarwal et~al.(2008)Agarwal, Snavely, and Seitz]{AgarwalSS08}
S.~Agarwal, N.~Snavely, and S.~M. Seitz.
\newblock Fast algorithms for $l_{\infty}$ problems in multiview geometry.
\newblock In \emph{{Proc. CVPR}}, pages 1--8, 2008.

\bibitem[Agarwal et~al.(2009)Agarwal, Snavely, Simon, Seitz, and
  Szeliski]{agarwal2009}
S.~Agarwal, N.~Snavely, I.~Simon, S.~M. Seitz, and R.~Szeliski.
\newblock Building rome in a day.
\newblock In \emph{{Proc. ICCV}}, 2009.

\bibitem[{Arie-Nachimson} et~al.(2012){Arie-Nachimson}, Kovalsky,
  {Kemelmacher-Shlizerman}, Singer, and Basri]{Arie2012}
M.~{Arie-Nachimson}, S.~Z. Kovalsky, I.~{Kemelmacher-Shlizerman}, A.~Singer,
  and R.~Basri.
\newblock Global motion estimation from point matches.
\newblock In \emph{Proc. 3DPVT}, 2012.

\bibitem[Boyd et~al.(2011)Boyd, Parikh, Chu, Peleato, and
  Eckstein]{Boyd2011ADMM}
S.~Boyd, N.~Parikh, E.~Chu, B.~Peleato, and J.~Eckstein.
\newblock Distributed optimization and statistical learning via the alternating
  direction method of multipliers.
\newblock \emph{Found. Trends Mach. Learn.}, 3\penalty0 (1):\penalty0 1--122,
  2011.
\newblock ISSN 1935-8237.

\bibitem[Brand et~al.(2004)Brand, Antone, and Teller]{Brand04s}
M.~Brand, M.~Antone, and S.~Teller.
\newblock Spectral solution of large-scale extrinsic camera calibration as a
  graph embedding problem.
\newblock In \emph{{Proc. ECCV}}, 2004.

\bibitem[Candes and Tao(2005)]{candes2005decoding}
E.~J. Candes and T.~Tao.
\newblock Decoding by linear programming.
\newblock \emph{Information Theory, IEEE Transactions on}, 51\penalty0
  (12):\penalty0 4203--4215, 2005.

\bibitem[Chatterjee and Govindu(2013)]{chatterjee2013efficient}
A.~Chatterjee and V.~M. Govindu.
\newblock Efficient and robust large-scale rotation averaging.
\newblock In \emph{{Proc. ICCV}}, pages 521--528, 2013.

\bibitem[Courchay et~al.(2010)Courchay, Dalalyan, Keriven, and
  Sturm]{CourchayDKS10}
J.~Courchay, A.~S. Dalalyan, R.~Keriven, and P.~Sturm.
\newblock Exploiting loops in the graph of trifocal tensors for calibrating a
  network of cameras.
\newblock In \emph{{Proc. ECCV}}, pages 85--99, 2010.

\bibitem[Crandall et~al.(2011)Crandall, Owens, Snavely, and
  Huttenlocher]{Crandall2011}
D.~Crandall, A.~Owens, N.~Snavely, and D.~P. Huttenlocher.
\newblock Discrete-continuous optimization for large-scale structure from
  motion.
\newblock In \emph{{Proc. CVPR}}, pages 3001--3008, 2011.

\bibitem[Dalalyan and Keriven(2009)]{Dalalyan09l1p}
A.~Dalalyan and R.~Keriven.
\newblock L1-penalized robust estimation for a class of inverse problems
  arising in multiview geometry.
\newblock In \emph{NIPS}, 2009.

\bibitem[Enqvist et~al.(2011)Enqvist, Kahl, and Olsson]{enqvist11}
O.~Enqvist, F.~Kahl, and C.~Olsson.
\newblock Non-sequential structure from motion.
\newblock In \emph{Workshop on Omnidirectional Vision, Camera Networks and
  Non-Classical Cameras}, 2011.

\bibitem[Ferraz et~al.(2014)Ferraz, Binefa, and Moreno-Noguer]{ferraz2014}
L.~Ferraz, X.~Binefa, and F.~Moreno-Noguer.
\newblock Very fast solution to the pnp problem with algebraic outlier
  rejection.
\newblock In \emph{{Proc. CVPR}}, pages 501--508, 2014.

\bibitem[Fitzgibbon and Zisserman(1998)]{fitzgibbon1998}
A.~Fitzgibbon and A.~Zisserman.
\newblock Automatic camera recovery for closed or open image sequences.
\newblock \emph{{Proc. ECCV}}, pages 311--326, 1998.

\bibitem[Goldstein and Osher(2009)]{goldstein2009split}
T.~Goldstein and S.~Osher.
\newblock The split bregman method for l1-regularized problems.
\newblock \emph{SIAM Journal on Imaging Sciences}, 2\penalty0 (2):\penalty0
  323--343, 2009.

\bibitem[Govindu(2001)]{Govindu01c}
V.~M. Govindu.
\newblock Combining two-view constraints for motion estimation.
\newblock In \emph{{Proc. CVPR}}, pages 218--225, 2001.

\bibitem[Govindu(2004)]{Govindu04}
V.~M. Govindu.
\newblock Lie-algebraic averaging for globally consistent motion estimation.
\newblock In \emph{{Proc. CVPR}}, 2004.

\bibitem[Govindu(2006)]{Govindu06}
V.~M. Govindu.
\newblock Robustness in motion averaging.
\newblock In \emph{{Proc. ACCV}}, 2006.

\bibitem[Haner and Heyden(2012)]{haner2012covariance}
S.~Haner and A.~Heyden.
\newblock Covariance propagation and next best view planning for 3d
  reconstruction.
\newblock In \emph{{Proc. ECCV}}, pages 545--556, 2012.

\bibitem[Hartley and Zisserman(2003)]{hartley2003book}
R.~Hartley and A.~Zisserman.
\newblock \emph{Multiple View Geometry in Computer Vision}.
\newblock Cambridge University Press, 2003.

\bibitem[Hartley et~al.(2013)Hartley, Trumpf, Dai, and Li]{hartley2013rotation}
R.~Hartley, J.~Trumpf, Y.~Dai, and H.~Li.
\newblock Rotation averaging.
\newblock \emph{{IJCV}}, pages 1--39, 2013.

\bibitem[Havlena et~al.(2009)Havlena, Torii, Knopp, and
  Pajdla]{havlena2009randomized}
M.~Havlena, A.~Torii, J.~Knopp, and T.~Pajdla.
\newblock Randomized structure from motion based on atomic 3d models from
  camera triplets.
\newblock In \emph{{Proc. CVPR}}, pages 2874--2881, 2009.

\bibitem[Jiang et~al.(2013)Jiang, Cui, and Tan]{jiang2013g}
N.~Jiang, Z.~Cui, and P.~Tan.
\newblock A global linear method for camera pose registration.
\newblock In \emph{{Proc. ICCV}}, 2013.

\bibitem[Kahl and Hartley(2007)]{Kahl2007m}
F.~Kahl and R.~Hartley.
\newblock Multiple view geometry under the $l_{\infty}$-norm.
\newblock \emph{{IEEE Trans. PAMI}}, 30:\penalty0 1603--1617, 2007.

\bibitem[Kneip et~al.(2011)Kneip, Chli, and Siegwart]{kneip2011robust}
L.~Kneip, M.~Chli, and R.~Siegwart.
\newblock Robust real-time visual odometry with a single camera and an imu.
\newblock In \emph{{Proc. BMVC}}, pages 1--11, 2011.

\bibitem[Lhuillier and Quan(2005)]{Lhuillier2005}
M.~Lhuillier and L.~Quan.
\newblock A quasi-dense approach to surface reconstruction from uncalibrated
  images.
\newblock \emph{{IEEE Trans. PAMI}}, 27\penalty0 (3):\penalty0 418--433, 2005.

\bibitem[Li(2009)]{Li2009efficient}
H.~Li.
\newblock Efficient reduction of l-infinity geometry problems.
\newblock In \emph{{Proc. CVPR}}, pages 2695--2702, 2009.

\bibitem[Li and Hartley(2006)]{Li2006}
H.~Li and R.~Hartley.
\newblock Five-point motion estimation made easy.
\newblock In \emph{Proc. ICPR}, pages 630--633, 2006.

\bibitem[Lin et~al.(2010)Lin, Chen, and Ma]{lin2010augmented}
Z.~Lin, M.~Chen, and Y.~Ma.
\newblock The augmented lagrange multiplier method for exact recovery of
  corrupted low-rank matrices.
\newblock \emph{arXiv preprint arXiv:1009.5055}, 2010.

\bibitem[Lin et~al.(2011)Lin, Liu, and Su]{lin2011linearized}
Z.~Lin, R.~Liu, and Z.~Su.
\newblock Linearized alternating direction method with adaptive penalty for
  low-rank representation.
\newblock In \emph{Advances in neural information processing systems}, pages
  612--620, 2011.

\bibitem[Martinec and Pajdla(2007)]{MartinecP07}
D.~Martinec and T.~Pajdla.
\newblock Robust rotation and translation estimation in multiview
  reconstruction.
\newblock In \emph{{Proc. CVPR}}, pages 1--8, 2007.

\bibitem[Moulon et~al.(2013)Moulon, Monasse, and Marlet]{moulon2013gf}
P.~Moulon, P.~Monasse, and R.~Marlet.
\newblock Global fusion of relative motions for robust, accurate and scalable
  structure from motion.
\newblock In \emph{{Proc. ICCV}}, 2013.

\bibitem[Nist\'{e}r(2004)]{nister04}
D.~Nist\'{e}r.
\newblock An efficient solution to the five-point relative pose problem.
\newblock \emph{{IEEE Trans. PAMI}}, 26:\penalty0 756--777, 2004.

\bibitem[Olsson et~al.(2007)Olsson, Eriksson, and Kahl]{Olsson07}
C.~Olsson, A.~Eriksson, and F.~Kahl.
\newblock Efficient optimization for $l_{\infty}$ problems using
  pseudoconvexity.
\newblock In \emph{{Proc. ICCV}}, 2007.

\bibitem[Olsson et~al.(2010)Olsson, Eriksson, and Hartley]{OlssonEH10}
C.~Olsson, A.~Eriksson, and R.~Hartley.
\newblock Outlier removal using duality.
\newblock In \emph{{Proc. CVPR}}, pages 1450--1457, 2010.

\bibitem[\"{O}zyesil and Singer(2015)]{Ozyesil2015}
O.~\"{O}zyesil and A.~Singer.
\newblock Robust camera location estimation by convex programming.
\newblock In \emph{{Proc. CVPR}}, 2015.

\bibitem[Parikh and Boyd(2013)]{parikh2013proximal}
N.~Parikh and S.~Boyd.
\newblock Proximal algorithms.
\newblock \emph{Foundations and Trends in Optimization}, 1\penalty0
  (3):\penalty0 123--231, 2013.

\bibitem[Pollefeys et~al.(2004)Pollefeys, {Van Gool}, Vergauwen, Verbiest,
  Cornelis, Tops, and Koch]{Pollefeys2004}
M.~Pollefeys, L.~{Van Gool}, M.~Vergauwen, F.~Verbiest, K.~Cornelis, J.~Tops,
  and R.~Koch.
\newblock Visual modeling with a hand-held camera.
\newblock \emph{{IJCV}}, 59:\penalty0 207--232, 2004.

\bibitem[Pollefeys et~al.(2008)Pollefeys, Nist{\'e}r, Frahm, Akbarzadeh,
  Mordohai, Clipp, Engels, Gallup, Kim, Merrell, et~al.]{pollefeys2008detailed}
M.~Pollefeys, D.~Nist{\'e}r, J-M Frahm, A.~Akbarzadeh, P.~Mordohai, B.~Clipp,
  C.~Engels, D.~Gallup, S-J Kim, P.~Merrell, et~al.
\newblock Detailed real-time urban 3d reconstruction from video.
\newblock \emph{{IJCV}}, 78\penalty0 (2-3):\penalty0 143--167, 2008.

\bibitem[Rother(2003)]{Rother2003}
C.~Rother.
\newblock \emph{Multi-View Reconstruction and Camera Recovery using a Real or
  Virtual Reference Plane}.
\newblock PhD thesis, January 2003.

\bibitem[Sim and Hartley(2006)]{Sim2006}
K.~Sim and R.~Hartley.
\newblock Recovering camera motion using $l_{\infty}$ minimization.
\newblock In \emph{{Proc. CVPR}}, 2006.

\bibitem[Sinha et~al.(2010)Sinha, Steedly, and Szeliski]{sinha2010ml}
S.~Sinha, D.~Steedly, and R.~Szeliski.
\newblock A multi-stage linear approach to structure from motion.
\newblock In \emph{ECCV Workshop on Reconstruction and Modeling of Large-Scale
  3D Virtual Environments}, 2010.

\bibitem[Snavely et~al.(2006)Snavely, Seitz, and Szeliski]{snavely2006}
N.~Snavely, S.~M. Seitz, and R.~Szeliski.
\newblock Photo tourism: exploring photo collections in 3d.
\newblock \emph{{ACM Trans. on Graph.}}, 25:\penalty0 835--846, 2006.

\bibitem[Snavely et~al.(2008)Snavely, Seitz, and Szeliski]{snavely2008modeling}
N.~Snavely, S.~M. Seitz, and R.~Szeliski.
\newblock Modeling the world from internet photo collections.
\newblock \emph{{IJCV}}, 80\penalty0 (2):\penalty0 189--210, 2008.

\bibitem[Strecha et~al.(2008)Strecha, {von Hansen}, {Van Gool}, Fua, and
  Thoennessen]{Strecha2008}
C.~Strecha, W.~{von Hansen}, L.~{Van Gool}, P.~Fua, and U.~Thoennessen.
\newblock On benchmarking camera calibration and multi-view stereo for high
  resolution imagery.
\newblock In \emph{{Proc. CVPR}}, 2008.

\bibitem[Wilson and Snavely(2014)]{Wilson2014}
K.~Wilson and N.~Snavely.
\newblock Robust global translations with 1dsfm.
\newblock In \emph{Proc. ECCV (3)}, pages 61--75, 2014.

\bibitem[Wu(2013)]{wu2013visualsfm}
C.~Wu.
\newblock Towards linear-time incremental structure from motion.
\newblock In \emph{{Proc. 3DV}}, 2013.

\bibitem[Zach et~al.(2010)Zach, Klopschitz, and Pollefeys]{Zach2010}
C.~Zach, M.~Klopschitz, and M.~Pollefeys.
\newblock Disambiguating visual relations using loop constraints.
\newblock In \emph{Proc. CVPR}, 2010.

\end{thebibliography}
\end{document}